\documentclass[final]{elsarticle}
\usepackage[letterpaper, margin=2cm]{geometry}
\usepackage[bookmarks=false,pdfstartview={FitH}]{hyperref}
\usepackage{url}
\usepackage{graphicx,color}
\usepackage{subfigure}
\usepackage{paralist}
\usepackage{amsmath, amsthm, amssymb}
\usepackage{multirow}
\usepackage{threeparttable}
\usepackage{longtable}
\usepackage{fancyhdr}

\newcommand{\cX}{{\bar {\mathcal X}}}
\newcommand{\bX}{{\mathcal X}}
\newcommand{\bx}{{\bf x}}
\newcommand{\bc}{{\bf c}}

\begin{document}

\title{A Comparative Study of Efficient Initialization Methods for the K-Means Clustering Algorithm}

\author{M.\ Emre Celebi}
\address{Department of Computer Science\\Louisiana State University, Shreveport, LA, USA\\
        \href{mailto:ecelebi@lsus.edu}{ecelebi@lsus.edu}}
\author{Hassan A.\ Kingravi}
\address{School of Electrical and Computer Engineering\\Georgia Institute of Technology, Atlanta, GA, USA\\
         \href{mailto:kingravi@gatech.edu}{kingravi@gatech.edu}}
\author{Patricio A.\ Vela}
\address{School of Electrical and Computer Engineering\\Georgia Institute of Technology, Atlanta, GA, USA\\
         \href{mailto:pvela@gatech.edu}{pvela@gatech.edu}}

\begin{abstract}
K-means is undoubtedly the most widely used partitional clustering algorithm. Unfortunately, due to its gradient descent nature, this algorithm is highly sensitive to the initial placement of the cluster centers. Numerous initialization methods have been proposed to address this problem. In this paper, we first present an overview of these methods with an emphasis on their computational efficiency. We then compare eight commonly used linear time complexity initialization methods on a large and diverse collection of data sets using various performance criteria. Finally, we analyze the experimental results using non-parametric statistical tests and provide recommendations for practitioners. We demonstrate that popular initialization methods often perform poorly and that there are in fact strong alternatives to these methods.
\end{abstract}

\maketitle

\section{Introduction}
\label{sec_intro}

Clustering, the unsupervised classification of patterns into groups, is one of the most important tasks in exploratory  data analysis \cite{Jain99}. Primary goals of clustering include gaining insight into data (detecting anomalies, identifying salient features, etc.), classifying data, and compressing data. Clustering has a long and rich history in a variety of scientific disciplines including anthropology, biology, medicine, psychology, statistics, mathematics, engineering, and computer science. As a result, numerous clustering algorithms have been proposed since the early 1950s \cite{Jain10}.
\par
Clustering algorithms can be broadly classified into two groups: hierarchical and partitional \cite{Jain10}. Hierarchical algorithms recursively find nested clusters either in a top-down (divisive) or bottom-up (agglomerative) fashion. In contrast, partitional  algorithms find all the clusters simultaneously as a partition of the data and do not impose a hierarchical structure. Most hierarchical algorithms have quadratic or higher complexity in the number of data points \cite{Jain99} and therefore are not suitable for large data sets, whereas partitional algorithms often have lower complexity.
\par
Given a data set $\bX = \{ \bx_1, \bx_2, \ldots, \bx_N \}$ in $\mathbb{R}^D$, i.e., $N$ points (vectors) each with $D$ attributes (components), hard partitional algorithms divide $\bX$ into $K$ exhaustive and mutually exclusive clusters $\mathcal{P} = \{ P_1, P_2, \ldots, P_K \},$ $\;\; \bigcup\nolimits_{i = 1}^K {P_i = \bX},$ $\;\; P_i \cap P_j = \emptyset$ for $1 \leq i \neq j \leq K$. These algorithms usually generate clusters by optimizing a criterion function. The most intuitive and frequently used criterion function is the Sum of Squared Error (\texttt{SSE}) given by:

\begin{equation}
\label{eq_sse}
 \mathtt{SSE} = \sum\limits_{i = 1}^K {\sum\limits_{\bx_j \in P_i} {\left\| {\bx_j - \bc_i } \right\|_2^2 } }
\end{equation}

\noindent where $\| . \|_2$ denotes the Euclidean ($\mathcal{L}_2$) norm and
$\bc_i =
{1 \mathord{\left/ {\vphantom {1 {\left| P_i \right|}}} \right.
 \kern-\nulldelimiterspace} {\left| P_i \right|}}\sum\nolimits_{\bx_j \in P_i} {\bx_j}$ is the centroid of cluster $P_i$ whose cardinality is $\left| P_i \right|$. The optimization of \eqref{eq_sse} is often referred to as the minimum \texttt{SSE} clustering (MSSC) problem.
\par
The number of ways in which a set of $N$ objects can be partitioned into $K$ non-empty groups is given by Stirling numbers of the second kind:

\begin{equation}
\label{eq_num_parts}
\mathcal{S}(N,K) =
\frac{1}
{{K!}}\sum\limits_{i = 0}^K {( - 1)^{K - i} \left( \begin{gathered}
  K \hfill \\
  i \hfill \\
\end{gathered} \right)} i^N
\end{equation}

\noindent which can be approximated by $K^N/K!$ It can be seen that a complete enumeration of all possible clusterings to determine the global minimum of \eqref{eq_sse} is clearly computationally prohibitive except for very small data sets \cite{Kaufman90}. In fact, this non-convex optimization problem is proven to be NP-hard even for $K = 2$ \cite{Aloise09} or $D = 2$ \cite{Mahajan12}. Consequently, various heuristics have been developed to provide approximate solutions to this problem \cite{Tarsitano03}. Among these heuristics, Lloyd's algorithm \cite{Lloyd82}, often referred to as the (batch) k-means algorithm, is the simplest and most commonly used one. This algorithm starts with $K$ arbitrary centers, typically chosen uniformly at random from the data points. Each point is assigned to the nearest center and then each center is recalculated as the mean of all points assigned to it. These two steps are repeated until a predefined termination criterion is met.
\par
The k-means algorithm is undoubtedly the most widely used partitional clustering algorithm \cite{Jain99, Jain10}. Its popularity can be attributed to several reasons. First, it is conceptually simple and easy to implement. Virtually every data mining software includes an implementation of it. Second, it is versatile, i.e., almost every aspect of the algorithm (initialization, distance function, termination criterion, etc.) can be modified. This is evidenced by hundreds of publications over the last fifty years that extend k-means in various ways. Third, it has a time complexity that is linear in $N$, $D$, and $K$ (in general, $D \ll N$ and $K \ll N$). For this reason, it can be used to initialize more expensive clustering algorithms such as expectation maximization \cite{Bradley98}, DBSCAN \cite{Dash01}, and spectral clustering \cite{Chen11}. Furthermore, numerous sequential \cite{Kanungo02, Hamerly10} and parallel \cite{Chen10} acceleration techniques are available in the literature. Fourth, it has a storage complexity that is linear in $N$, $D$, and $K$. In addition, there exist disk-based variants that do not require all points to be stored in memory \cite{Ordonez04}. Fifth, it is guaranteed to converge \cite{Selim84} at a quadratic rate \cite{Bottou95}. Finally, it is invariant to data ordering, i.e., random shufflings of the data points.
\par
On the other hand, k-means has several significant disadvantages. First, it requires the number of clusters, $K$, to be specified \emph{a priori}. The value of this parameter can be determined automatically by means of various cluster validity measures \cite{Vendramin10}. Second, it can only detect compact, hyperspherical clusters that are well separated. This can be alleviated by using a more general distance function such as the Mahalanobis distance, which permits the detection of hyperellipsoidal clusters \cite{Mao96}. Third, due its utilization of the squared Euclidean distance, it is sensitive to noise and outlier points since even a few such points can significantly influence the means of their respective clusters. This can addressed by outlier pruning \cite{Zhang03} or using a more robust distance function such as City-block ($\mathcal{L}_1$) distance. Fourth, due to its gradient descent nature, it often converges to a local minimum of the criterion function \cite{Selim84}. For the same reason, it is highly sensitive to the selection of the initial centers. Adverse effects of improper initialization include empty clusters, slower convergence, and a higher chance of getting stuck in bad local minima \cite{Celebi11}. Fortunately, all of these drawbacks except the first one can be remedied by using an adaptive initialization method (IM).
\par
In this study, we investigate some of the most popular IMs developed for the k-means algorithm. Our motivation is three-fold. First, a large number of IMs have been proposed in the literature and thus a systematic study that reviews and compares these methods is desirable. Second, these IMs can be used to initialize other partitional clustering algorithms such as fuzzy c-means and its variants and expectation maximization. Third, most of these IMs can be used independently of k-means as standalone clustering algorithms.
\par
This study differs from earlier studies of a similar nature \cite{Pena99, He04} in several respects:
\begin{inparaenum}[(i)]
 \item a more comprehensive overview of the existing IMs is provided,
 \item the experiments involve a larger set of methods and a significantly more diverse collection of data sets,
 \item in addition to clustering effectiveness, computational efficiency is used as a performance criterion, and
 \item the experimental results are analyzed more thoroughly using non-parametric statistical tests.
\end{inparaenum}

\par
The rest of the paper is organized as follows. Section \ref{sec_init_methods} presents a survey of k-means IMs. Section \ref{sec_exp_setup} describes the experimental setup. Section \ref{sec_exp_results} presents the experimental results, while Section \ref{sec_conc} gives the conclusions.

\section{Initialization Methods for K-Means}
\label{sec_init_methods}

In this section, we briefly review some of the commonly used IMs with an emphasis on their time complexity (with respect to $N$). In each complexity class, methods are presented in chronologically ascending order.

\subsection{Linear Time-Complexity Initialization Methods}
Forgy's method \cite{Forgy65} assigns each point to one of the $K$ clusters uniformly at random. The centers are then given by the centroids of these initial clusters. This method has no theoretical basis, as such random clusters have no internal homogeneity \cite{Anderberg73}.
\par
Jancey's method \cite{Jancey66} assigns to each center a synthetic point randomly generated within the data space. Unless the data set fills the space, some of these centers may be quite distant from any of the points \cite{Anderberg73}, which might lead to the formation of empty clusters.
\par
MacQueen \cite{MacQueen67} proposed two different methods. The first one, which is the default option in the Quick Cluster procedure of IBM SPSS Statistics \cite{Norusis11}, takes the first $K$ points in $\bX$ as the centers. An obvious drawback of this method is its sensitivity to data ordering. The second method chooses the centers randomly from the data points. The rationale behind this method is that random selection is likely to pick points from dense regions, i.e., points that are good candidates to be centers. However, there is no mechanism to avoid choosing outliers or points that are too close to each other \cite{Anderberg73}. Multiple runs of this method is the standard way of initializing k-means \cite{Bradley98}. It should be noted that this second method is often mistakenly attributed to Forgy \cite{Forgy65}.
\par
Ball and Hall's method \cite{Ball67} takes the centroid of $\bX$, i.e., $\cX = {1 \mathord{\left/ {\vphantom {1 N}} \right. \kern-\nulldelimiterspace} N}\sum\nolimits_{j = 1}^N {\bx_j}$, as the first center. It then traverses the points in arbitrary order and takes a point as a center if it is at least $T$ units apart from the previously selected centers until $K$ centers are obtained. The purpose of the distance threshold $T$ is to ensure that the seed points are well separated. However, it is difficult to decide on an appropriate value for $T$. In addition, the method is sensitive to data ordering.
\par
The Simple Cluster Seeking method \cite{Tou74} is identical to Ball and Hall's method with the exception that the first point in $\bX$ is taken as the first center. This method is used in the FASTCLUS procedure of SAS \cite{SAS09}.
\par
Sp\"{a}th's method \cite{Spath77} is similar to Forgy's method with the exception that the points are assigned to the clusters in a cyclical fashion, i.e., the $j$-th ($j \in \{ 1, 2, \ldots, N \}$) point is assigned to the $(j - 1 \pmod{K} + 1 )$-th cluster. In contrast to Forgy's method, this method is sensitive to data ordering.
\par
Maximin method \cite{Gonzalez85, Katsavounidis94} chooses the first center $\bc_1$ arbitrarily and the $i$-th ($i \in \{2, 3, \ldots, K\}$) center $\bc_i$ is chosen to be the point that has the greatest minimum-distance to the previously selected centers, i.e., $\bc_1, \bc_2, \ldots, \bc_{i-1}$. This method was originally developed as a $2$-approximation to the $K$-center clustering problem\footnote{Given a set of $N$ points in a metric space, the goal of $K$-center clustering is to find $K$ representative points (centers) such that the maximum distance of a point to a center is minimized.}. It should be noted that, motivated by a vector quantization application, Katsavounidis \emph{et al.}'s variant \cite{Katsavounidis94} takes the point with the greatest Euclidean norm as the first center.
\par
Al-Daoud's density-based method \cite{AlDaoud96} first uniformly partitions the data space into $M$ disjoint hypercubes. It then randomly selects $K N_m/N$ points from hypercube $m$ ($m \in \{ 1, 2, \ldots, M \}$) to obtain a total of $K$ centers ($N_m$ is the number of points in hypercube $m$). There are two main disadvantages associated with this method. First, it is difficult to decide on an appropriate value for $M$. Second, the method has a storage complexity of $\mathcal{O}(2^{BD})$, where $B$ is the number of bits allocated to each attribute.
\par
Bradley and Fayyad's method \cite{Bradley98} starts by randomly partitioning the data set into $J$ subsets. These subsets are clustered using k-means initialized by MacQueen's second method producing $J$ sets of intermediate centers each with $K$ points. These center sets are combined into a superset, which is then clustered by k-means $J$ times, each time initialized with a different center set. Members of the center set that give the least \texttt{SSE} are then taken as the final centers.
\par
Pizzuti \emph{et al.} \cite{Pizzuti99} improved upon Al-Daoud's density-based method using a multiresolution grid approach. Their method starts with $2^D$ hypercubes and iteratively splits these as the number of points they receive increases. Once the splitting phase is completed, the centers are chosen from the densest hypercubes.
\par
The k-means++ method \cite{Arthur07} interpolates between MacQueen's second method and the maximin method. It chooses the first center randomly and the $i$-th ($i \in \{ 2, 3, \ldots, K \}$) center is chosen to be $\bx' \in \bX$ with a probability of $\frac{{md\left( {\bx'} \right)^2 }}{{\sum\nolimits_{j = 1}^N {md(\bx_j)^2 }}}$, where $md(\bx)$ denotes the minimum-distance from a point $\bx$ to the previously selected centers. This method yields an $\Theta(\log{K})$ approximation to the MSSC problem. The greedy k-means++ method probabilistically selects $\log(K)$ centers in each round and then greedily selects the center that most reduces the \texttt{SSE}. This modification aims to avoid the unlikely event of choosing two centers that are close to each other.
\par
The PCA-Part method \cite{Su07} uses a divisive hierarchical approach based on PCA (Principal Component Analysis) \cite{Hotelling36}. Starting from an initial cluster that contains the entire data set, the method iteratively selects the cluster with the greatest \texttt{SSE} and divides it into two subclusters using a hyperplane that passes through the cluster centroid and is orthogonal to the principal eigenvector of the cluster covariance matrix. This procedure is repeated until $K$ clusters are obtained. The centers are then given by the centroids of these clusters. The Var-Part method \cite{Su07} is an approximation to PCA-Part, where the covariance matrix of the cluster to be split is assumed to be diagonal. In this case, the splitting hyperplane is orthogonal to the coordinate axis with the greatest variance.
\par
Lu \emph{et al.}'s method \cite{Lu08} uses a two-phase pyramidal approach. The attributes of each point are first encoded as integers using $2^Q$-level quantization, where $Q$ is a resolution parameter. These integer points are considered to be at level $0$ of the pyramid. In the bottom-up phase, starting from level $0$, neighboring data points at level $k$ ($k \in \{ 0, 1, \ldots \}$) are averaged to obtain weighted points at level $k+1$ until at least $20K$ points are obtained. Data points at the highest level are refined using k-means initialized with the $K$ points with the largest weights. In the top-down phase, starting from the highest level, centers at level $k + 1$ are projected onto level $k$ and then used to initialize the $k$-th level clustering. The top-down phase terminates when level $0$ is reached. The centers at this level are then inverse quantized to obtain the final centers. The performance of this method degrades with increasing dimensionality \cite{Lu08}.
\par
Onoda \emph{et al.}'s method \cite{Onoda12} first calculates $K$ Independent Components (ICs) \cite{Hyvarinen99} of $\bX$ and then chooses the $i$-th ($i \in \{ 1, 2, \ldots, K \}$) center as the point that has the least cosine distance from the $i$-th IC.

\subsection{Loglinear Time-Complexity Initialization Methods}
Hartigan's method \cite{Hartigan79} first sorts the points according to their distances to $\cX$. The $i$-th ($i \in \{ 1, 2, \ldots, K \}$) center is then chosen to be the ($1 + ( i - 1 )N/K$)-th point. This method is an improvement over MacQueen's first method in that it is invariant to data ordering and is more likely to produce seeds that are well separated. The computational cost of this method is dominated by the complexity of sorting, which is $\mathcal{O}(N \log{N})$.
\par
Al-Daoud's variance-based method \cite{AlDaoud05} first sorts the points on the attribute with the greatest variance and then partitions them into $K$ groups along the same dimension. The centers are then chosen to be the points that correspond to the medians of these groups. Note that this method disregards all attributes but one and therefore is likely to be effective only for data sets in which the variability is mostly on one dimension.
\par
Redmond and Heneghan's method \cite{Redmond07} first constructs a kd-tree of the data points to perform density estimation and then uses a modified maximin method to select $K$ centers from densely populated leaf buckets. The computational cost of this method is dominated by the complexity of kd-tree construction, which is $\mathcal{O}(N \log{N})$.
\par
The ROBIN (ROBust INitialization) method \cite{AlHasan09} uses a local outlier factor (LOF) \cite{Breunig00} to avoid selecting outlier points as centers. In iteration $i$ ($i \in \{1,2, \ldots, K\}$), the method first sorts the data points in decreasing order of their minimum-distance to the previously selected centers. It then traverses the points in sorted order and selects the first point that has an LOF value close to $1$ as the $i$-th center. The computational cost of this method is dominated by the complexity of sorting, which is $\mathcal{O}(N \log{N})$.

\subsection{Quadratic-Complexity Initialization Methods}
Astrahan's method \cite{Astrahan70} uses two distance thresholds $d_1$ and $d_2$. It first calculates the \emph{density} of each point as the number of points within a distance of $d_1$. The points are sorted in decreasing order by their densities and the highest density point is chosen as the first center. Subsequent centers are chosen in order of decreasing density subject to the condition that each new center be at least at a distance of $d_2$ from the previously selected centers. This procedure is continued until no more centers can be chosen. Finally, if more than $K$ centers are chosen, hierarchical clustering is used to group the centers until only $K$ of them remain. The main problem with this method is that it is very sensitive to the values of $d_1$ and $d_2$. For example, if $d_1$ is too small there may be many isolated points with zero density whereas if it is too large a few centers will cover the entire data set \cite{Anderberg73}.
\par
Lance and Williams \cite{Lance67} suggested that the output of a hierarchical clustering algorithm can be used to initialize k-means. Despite the fact that such algorithms often have quadratic or higher complexity, this method is highly recommended in the statistics literature \cite{Milligan80a} possibly due to the limited size of the data sets in this field.
\par
Kaufman and Rousseeuw's method \cite{Kaufman90} takes $\cX$ as the first center and the $i$-th ($i \in \{2, 3, \ldots, K\}$) center is chosen to be the point that most reduces the \texttt{SSE}. Since pairwise distances between the data points need to be calculated in each iteration, the time complexity of this method is $\mathcal{O}(N^2)$.
\par
Cao \emph{et al.} \cite{Cao09} formalized Astrahan's density-based method within the framework of a neighborhood-based rough set model. In this model, the $\varepsilon$-neighborhood of a point is defined as the set of points within $\varepsilon$ distance from it according to a particular distance measure. Based on this neighborhood model, the concepts of \textit{cohesion} and \textit{coupling} are defined. The former is a measure of the centrality of a point with respect to its neighborhood; whereas the latter is a measure of separation between two neighborhoods. The method first sorts the data points in decreasing order of their \textit{cohesion} and takes the point with the greatest \textit{cohesion} as the first center. It then traverses the points in sorted order and takes the first point that has a \textit{coupling} of less than $\varepsilon$ with the previously selected centers as the $i$-th ($i \in \{2, 3, \ldots, K\}$) center. The computational cost of this method is dominated by the complexity of the $\varepsilon$-neighborhood calculations, which is $\mathcal{O}(N^2)$.

\subsection{Other Initialization Methods}
The binary-splitting method \cite{Linde80} takes $\cX$ as the first center. In iteration $t$ ($t \in \{ 1, 2, \ldots, \log_2K \}$), each of the existing $2^{t-1}$ centers is split into two new centers by subtracting and adding a fixed perturbation vector $\boldsymbol \epsilon$, i.e., $\bc_i - \boldsymbol \epsilon$ and $\bc_i + \boldsymbol \epsilon$ ($i \in \{ 1, 2, \ldots, 2^{t-1} \}$). These $2^t$ new centers are then refined using k-means. There are two main disadvantages associated with this method. First, there is no guidance on the selection of a proper value for $\boldsymbol \epsilon$, which determines the direction of the split \cite{Huang93}. Second, the method is computationally demanding since after each iteration k-means has to be run for the entire data set.
\par
The directed-search binary-splitting method \cite{Huang93} is an improvement over the binary-splitting method in that it determines the value of $\boldsymbol \epsilon$ using PCA. However, it has even higher computational requirements due to the calculation of the principal eigenvector in each iteration.
\par
The global k-means method \cite{Likas03} takes $\cX$ as the first center. In iteration $i$ ($i \in \{ 1, 2,  \ldots, K - 1 \}$) it considers each of the $N$ points in turn as a candidate for the $(i + 1)$-st center and runs k-means with $i + 1$ centers on the entire data set. This method is computationally prohibitive for large data sets as it involves $N (K - 1)$ runs of k-means on the entire data set.
\par
It should be noted that the two splitting methods and the global k-means method are not initialization methods \emph{per se}. These methods can be considered as complete clustering methods that utilize k-means as a local search procedure. For this reason, to the best of our knowledge, none of the initialization studies to date included these methods in their comparisons.
\par
We should also mention IMs based on metaheuristics such as simulated annealing \cite{Babu94a} and genetic algorithms \cite{Babu93}. These algorithms start from a random initial configuration (population) and use k-means to evaluate their solutions in each iteration (generation). There are two main disadvantages associated with these methods. First, they involve numerous parameters that are difficult to tune (initial temperature, cooling schedule, population size, crossover/mutation probability, etc.) \cite{Jain99}. Second, due to the large search space, they often require a large number of iterations, which renders them computationally prohibitive for all but the smallest data sets. Interestingly, with the recent developments in combinatorial optimization algorithms, it is now feasible to obtain globally minimum \texttt{SSE} clusterings for small data sets without resorting to metaheuristics \cite{Aloise10}.

\subsection{Linear vs.\ Superlinear Initialization Methods}
\label{sub_sec_linear_vs_super}

Based on the descriptions given above, it can be seen that superlinear methods often have more elaborate designs when compared to linear ones. An interesting feature of the superlinear methods is that they are often deterministic, which can be considered as an advantage especially when dealing with large data sets. In contrast, linear methods are often non-deterministic and/or order-sensitive. As a result, it is common practice to perform multiple runs of such methods and take the output of the run that produces the least \texttt{SSE} \cite{Bradley98}.
\par
A frequently cited advantage of the more elaborate methods is that they often lead to faster k-means convergence, i.e., require fewer iterations, and as a result the time gained during the clustering phase can offset the time lost during the initialization phase \cite{Su07, Redmond07, AlHasan09}. This may be true when a standard implementation of k-means is used. However, convergence speed may not be as important when a fast k-means variant is used as such methods often require significantly less time compared to a standard k-means implementation. In this study, we utilize a fast k-means variant based on triangle inequality \cite{Huang90} and partial distance elimination \cite{Bei85} techniques. As will be seen in \S \ref{sec_exp_results}, this fast and exact k-means implementation will diminish the computational efficiency differences among various IMs. In other words, we will demonstrate that elaborate methods that lead to faster k-means convergence are not necessarily more efficient than simple methods with slower convergence.

\section{Experimental Setup}
\label{sec_exp_setup}

\subsection{Data Set Descriptions}
In order to obtain a comprehensive evaluation of various IMs, we conducted two sets of experiments. The first experiment involved $32$ commonly used real data sets with sizes ranging from $214$ to $1,904,711$ points. Most of these data sets were obtained from the UCI Machine Learning Repository \cite{Frank11} (see Table \ref{tab_data_set}.) The second experiment involved a large number of synthetic data sets with varying clustering complexity. We used a recent algorithm proposed by Maitra and Melnykov \cite{Maitra10} to generate these data sets. This algorithm involves the calculation of the exact overlap ($\omega$) between each cluster pair, measured in terms of their total probability of misclassification, and guided simulation of Gaussian mixture components satisfying prespecified overlap characteristics. The algorithm was used with the following parameters: mean overlap ($\bar\omega \in \{ 0.025, 0.05, 0.1, 0.2 \}$), number of points ($N \in \{ 1024, 4096, 16384, 65536 \}$), number of attributes ($D \in \{ 2, 4, 8, 16, 32, 64 \}$), and number of classes ($K' \in \{ 2, 4, 6, 8, 10, 12 \}$).
\par
\begin{figure}[!ht]
\centering
 \subfigure[Easy ($\Omega = .103$)]{\includegraphics[width=0.32\columnwidth]{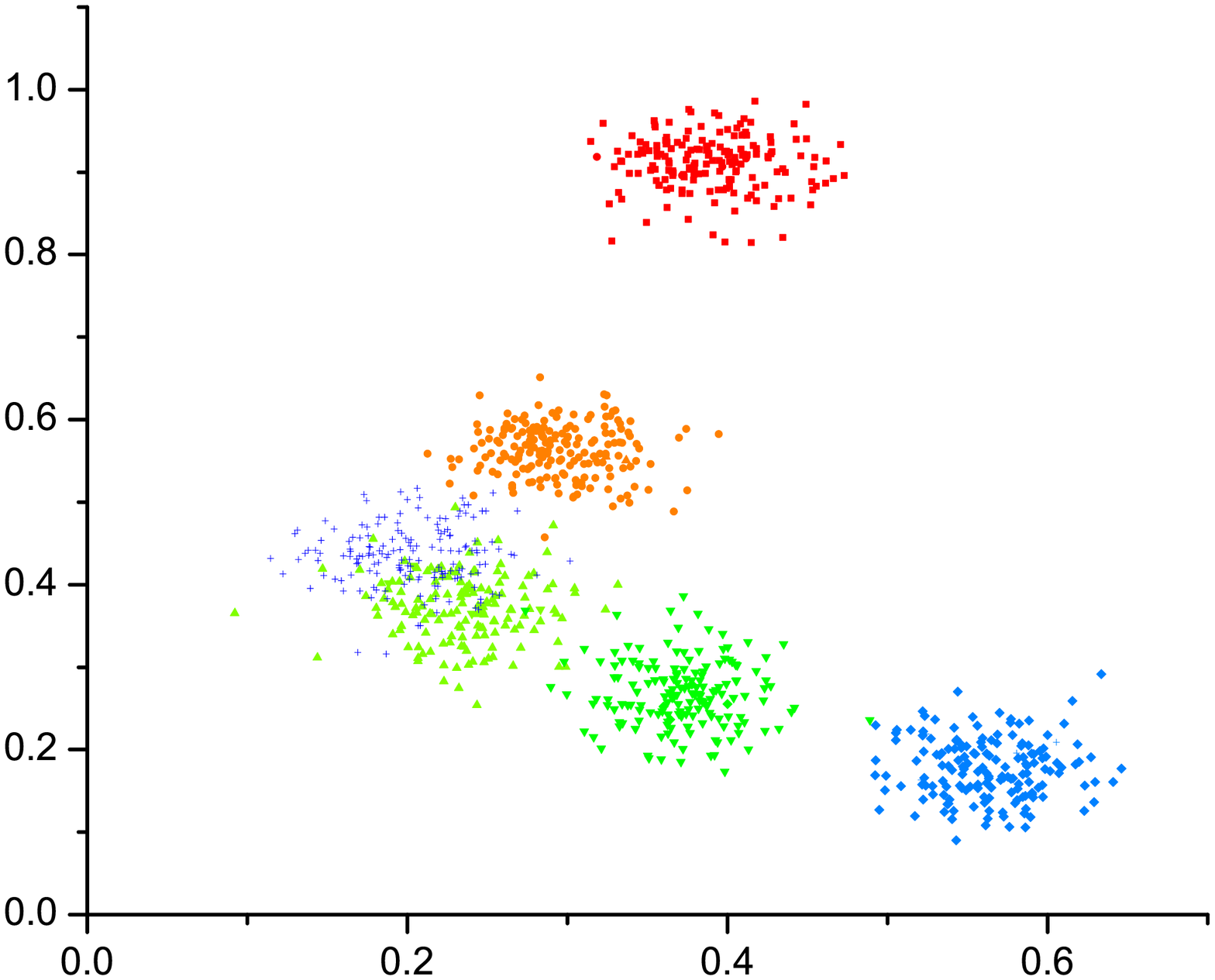}}
 \subfigure[Moderate ($\Omega = .369$)]{\includegraphics[width=0.32\columnwidth]{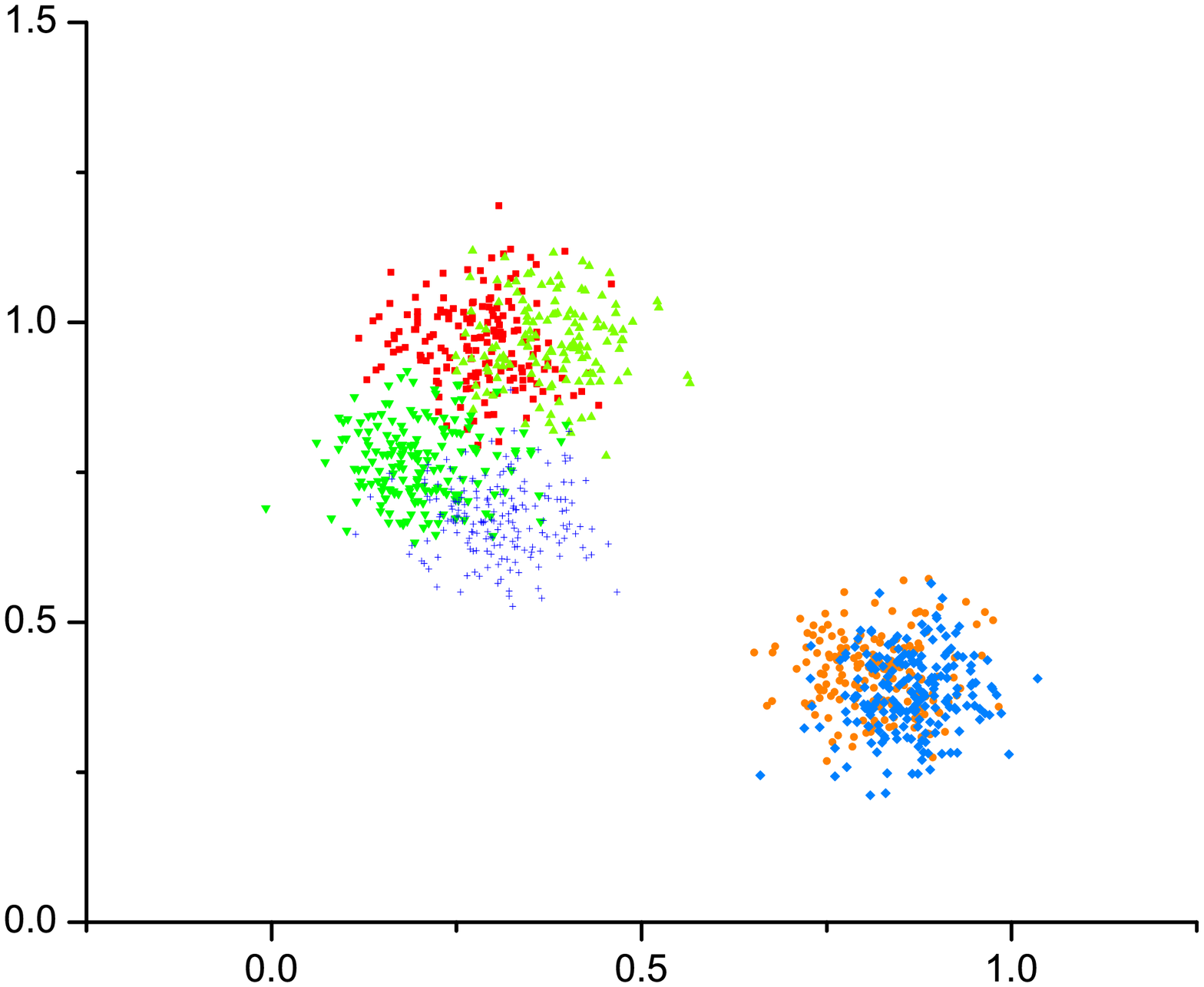}}
 \subfigure[Difficult ($\Omega = .569$)]{\includegraphics[width=0.32\columnwidth]{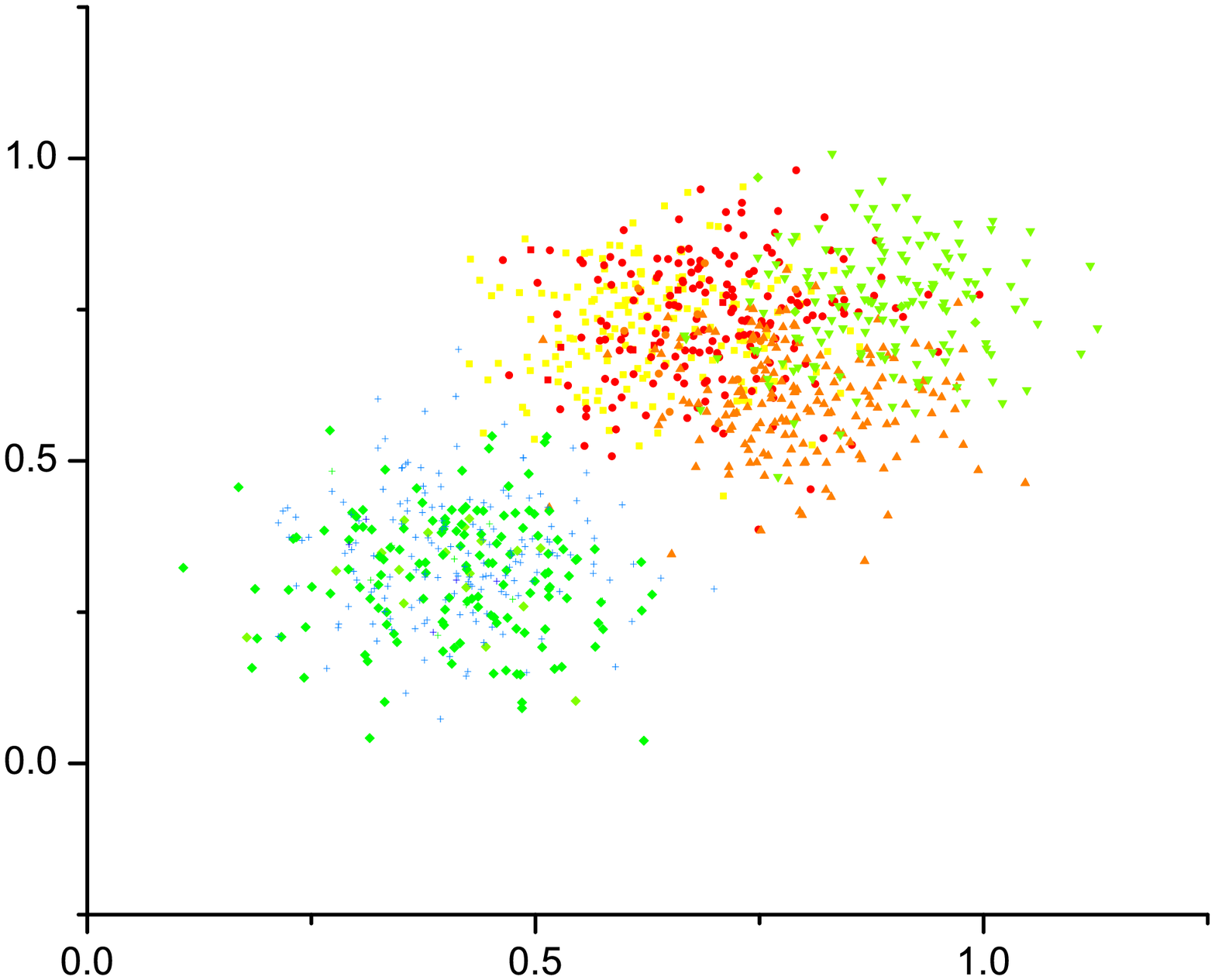}}
 \caption{Synthetic data sets with $K=6$ clusters}
 \label{fig_synthetic_data_sets}
\end{figure}

The parameter $\bar\omega$ denotes the mean overlap between pairs of clusters. However, we observed that two synthetic data sets with the same $\bar\omega$ can have considerably different clustering complexity. Therefore, we quantified clustering complexity using the following indirect approach. For each data set, we executed the k-means algorithm initialized with the ``true'' centers given by the cluster generation algorithm and calculated the \texttt{RAND}, \texttt{VD}, and \texttt{VI} measures (see \S \ref{sec_perf_crit}) upon convergence. The average of these measures, $\Omega$, was taken as a quantitative indicator of clustering complexity. Note that each of these normalized measures takes values from the $[0, 1]$ interval. For \texttt{RAND} larger values are better, whereas for \texttt{VD} and \texttt{VI} smaller values are better. Therefore, we inverted the \texttt{RAND} values by subtracting them from $1$ to make this measure compatible with the other two. Finally, using the aforementioned complexity quantification scheme, we generated $4,096$ synthetic data sets from each of the following complexity classes: easy ($0 \leq \Omega \leq 0.25$), moderate ($0.25 < \Omega \leq 0.5$), and difficult ($0.5 < \Omega \leq 1$). The total number of synthetic data sets was thus $3 \times 4,096 = 12,288$. Figure \ref{fig_synthetic_data_sets} shows sample data sets with $K=6$ clusters from each complexity class.

{
\small
\begin{table}[ht]
\linespread{1}
\centering
\caption{ \label{tab_data_set} Descriptions of Real Data Sets}
\begin{threeparttable}
\begin{tabular}{llrrr}
\hline
ID & Data Set & \# Points ($N$) & \# Attributes ($D$) & \# Classes ($K'$)\\
\hline
\hline
1 & Breast Cancer Wisconsin (Original) & 683 & 9 & 2\\
2 & Cloud Cover (DB1) & 1,024 & 10 & 8\tnote{\dag}\\
3 & Concrete Compressive Strength & 1,030 & 9 & 8\tnote{\dag}\\
4 & Corel Image Features & 68,040 & 25 & 16\tnote{\dag}\\
5 & Covertype & 581,012 & 10 & 7\\
6 & Ecoli & 336 & 7 & 8\\
7 & Steel Plates Faults & 1,941 & 27 & 7\\
8 & Glass Identification & 214 & 9 & 6\\
9 & Heart Disease & 297 & 13 & 5\\
10 & Ionosphere & 351 & 34 & 2\\
11 & ISOLET & 7,797 & 617 & 26\\
12 & Landsat Satellite (Statlog) & 6,435 & 36 & 6\\
13 & Letter Recognition & 20,000 & 16 & 26\\
14 & MAGIC Gamma Telescope & 19,020 & 10 & 2\\
15 & Multiple Features (Fourier) & 2,000 & 76 & 10\\
16 & MiniBooNE Particle Identification & 130,064 & 50 & 2\\
17 & Musk (Clean2) & 6598 & 166 & 2\\
18 & Optical Digits & 5,620 & 64 & 10\\
19 & Page Blocks Classification & 5,473 & 10 & 5\\
20 & Parkinsons & 5,875 & 18 & 42\tnote{\dag}\\
21 & Pen Digits & 10,992 & 16 & 10\\
22 & Person Activity & 164,860 & 3 & 11\\
23 & Pima Indians Diabetes & 768 & 8 & 2\\
24 & Image Segmentation & 2,310 & 19 & 7\\
25 & Shuttle (Statlog) & 58,000 & 9 & 7\\
26 & SPECTF Heart & 267 & 44 & 2\\
27 & Telugu Vowels \cite{Pal77} & 871 & 3 & 6\\
28 & Vehicle Silhouettes (Statlog) & 846 & 18 & 4\\
29 & Wall-Following Robot Navigation & 5,456 & 24 & 4\\
30 & Wine Quality & 6,497 & 11 & 7\\
31 & World TSP \cite{Cook11} & 1,904,711 & 2 & 7\tnote{\dag}\\
32 & Yeast & 1,484 & 8 & 10\\
\hline
\end{tabular}
\begin{tablenotes}
\item [\dag] Due to the unavailability of class labels, for data sets \#2, \#3, and \#4, $K'$ was chosen arbitrarily, whereas for \#20 and \#31, it was determined based on domain knowledge.
\end{tablenotes}
\end{threeparttable}
\end{table}
}

\subsection{Attribute Normalization}
In clustering tasks, normalization is a common preprocessing step that is necessary to prevent attributes with large ranges from dominating the distance calculations and also to avoid numerical instabilities in the computations. Two commonly used normalization schemes are linear scaling to unit range (min-max normalization) and linear scaling to unit variance (z-score normalization). Several studies revealed that the former scheme is preferable to the latter since the latter is likely to eliminate valuable between-cluster variation \cite{Milligan88, Su07}. As a result, we used min-max normalization to map the attributes of each real data set to the $[0,1]$ interval. Note that attributes of the synthetic data sets were already normalized by the cluster generation algorithm.

\subsection{Performance Criteria}
\label{sec_perf_crit}
The performance of the IMs was measured using five effectiveness (quality) and two efficiency (speed) criteria:

\begin{itemize}
\renewcommand{\labelitemi}{$\triangleright$}
 \item Initial \texttt{SSE}: This is the \texttt{SSE} value calculated after the initialization phase, before the clustering phase. It gives us a measure of the effectiveness of an IM by itself.
 \item Final \texttt{SSE}: This is the \texttt{SSE} value calculated after the clustering phase. It gives us a measure of the effectiveness of an IM when its output is refined by k-means. Note that this is the objective function of the k-means algorithm, i.e., \eqref{eq_sse}.
 \item Normalized Rand (\texttt{RAND}) \cite{Hubert85}, van Dongen (\texttt{VD}) \cite{vanDongen00}, and Variation of Information \cite{Meila07} criteria (\texttt{VI}): These are external validity measures that quantify the extent to which the clustering structure discovered by a clustering algorithm matches some external structure, e.g., one specified by the given class labels \cite{Wu09a, Wu09c}. In a recent comprehensive study, these three measures were found to be the best among 16 external validity measures \cite{Wu09a}. Note that each of these normalized measures takes values from the $[0, 1]$ interval.
 \item Number of Iterations: This is the number of iterations that k-means requires until reaching convergence when initialized by a particular IM. It is an efficiency measure independent of programming language, implementation style, compiler, and CPU architecture.
 \item CPU Time: This is the total CPU time taken by the initialization and clustering phases. This criterion is reported only for the real data sets.
\end{itemize}

All of the methods were implemented in the \texttt{C} language, compiled with the \texttt{gcc} v4.4.3 compiler, and executed on an Intel Xeon E5520 2.26GHz machine. Time measurements were performed using the \texttt{getrusage} function, which is capable of measuring CPU time to an accuracy of a microsecond. The MT19937 variant of the Mersenne Twister algorithm was used to generate high-quality pseudorandom numbers \cite{Matsumoto98}.

\par
The convergence of k-means was controlled by the disjunction of two criteria: the number of iterations reaches a maximum of $100$ or the relative improvement in \texttt{SSE} between two consecutive iterations drops below a threshold, i.e., $ {{\left( {\mathtt{SSE}_{i - 1} - \mathtt{SSE}_i} \right)} \mathord{\left/ {\vphantom {{\left( {\mathtt{SSE}_{i - 1} - \mathtt{SSE}_i} \right)} {\mathtt{SSE}_i}}} \right. \kern-\nulldelimiterspace} {\mathtt{SSE}_i}} \leq \varepsilon $, where $\mathtt{SSE}_i$ denotes the \texttt{SSE} value at the end of the $i$-th ($i \in \{ 1, 2, \ldots, 100 \}$) iteration. The convergence threshold was set to $\varepsilon = 10^{-6}$.

\section{Experimental Results and Discussion}
\label{sec_exp_results}

In this study, we focus on IMs that have time complexity linear in $N$. This is because k-means itself has linear complexity, which is perhaps the most important reason for its popularity. Therefore, an IM for k-means should not diminish this advantage of the algorithm. Eight commonly used, order-invariant IMs were included in the experiments: Forgy's method (\textsf{F}), MacQueen's second method (\textsf{M}), maximin (\textsf{X}), Bradley and Fayyad's method (\textsf{B}) with $J=10$, k-means++ (\textsf{K}), greedy k-means++ (\textsf{G}), Var-Part (\textsf{V}), and PCA-Part (\textsf{P}). It should be noted that among these methods only \textsf{V} and \textsf{P} are deterministic.
\par
In the experiments, each non-deterministic method was executed a $100$ times and statistics such as minimum, mean, and standard deviation were collected for the effectiveness criteria. In each run, the number of clusters ($K$) was set equal to the number of classes ($K'$), as commonly seen in the related literature \cite{Bradley98, Pizzuti99, He04, Arthur07, Su07, AlHasan09, Cao09}.
\par
Tables \ref{tab_real_final_sse} and \ref{tab_real_cpu_time} give the Final \texttt{SSE} and CPU time (in milliseconds) results for the real data sets, respectively. Note that, due to space limitations, only mean values are reported for the CPU time criterion. In order to determine if there are any statistically significant differences among the methods, we employed two non-parametric statistical tests \cite{Garcia08}: the Friedman test \cite{Friedman37} and Iman \& Davenport test \cite{Iman80}. These tests are alternatives to the parametric two-way analysis of variance (ANOVA) test. Their advantage over ANOVA is that they do not require normality or homoscedasticity, assumptions that are often violated in machine learning studies \cite{Luengo09, Garcia09}.
\par
Given $B$ blocks (subjects) and $T$ treatments (measurements), the null hypothesis ($H_0$) of the Friedman test is that populations within a block are identical. The alternative hypothesis ($H_1$) is that at least one treatment tends to yield larger (or smaller) values than at least one other treatment. The test statistic is calculated as follows \cite{Daniel00}. In the first step, the observations within each block are ranked separately, so each block contains a separate set of $T$ ranks. If ties occur, the tied observations are given the mean of the rank positions for which they are tied. If $H_0$ is true, the ranks in each block should be randomly distributed over the columns (treatments). Otherwise, we expect a lack of randomness in this distribution. For example, if a particular treatment is better than the others, we expect large (or small) ranks to `favor' that column. In the second step, the ranks in each column are summed. If $H_0$ is true, we expect the sums to be fairly close --- so close that we can attribute differences to chance. Otherwise, we expect to see at least one difference between pairs of rank sums so large that we cannot reasonably attribute it to sampling variability. The test statistic is given as:

\begin{equation}
\label{eq_friedman}
\chi_r ^2  = \frac{{12}}
{{BT(T + 1)}}\sum\limits_{j = 1}^T {R_j ^2} - 3B(T + 1)
\end{equation}

\noindent where $R_j$ ($j \in \{ 1, 2, \ldots, T \}$) is the rank sum of the $j$-th column. $\chi_r^2$ is approximately chi-square with $T-1$ degrees of freedom. $H_0$ is rejected at the $\alpha$ level of significance if the value of \eqref{eq_friedman} is greater than or equal to the critical chi-square value for $T-1$ degrees of freedom. Iman and Davenport \cite{Iman80} proposed the following statistic:

\begin{equation}
\label{eq_iman_davenport}
F_r  = \frac{{(B - 1)\chi _r ^2 }}
{{B(T - 1) - \chi _r ^2 }}
\end{equation}

\noindent which is distributed according to the F-distribution with $T - 1$ and $(T - 1)(B - 1)$ degrees of freedom. When compared to $\chi_r^2$, this statistic is not only less conservative, but also more accurate for small sample sizes \cite{Iman80}.
\par
In this study, blocks and treatments correspond to data sets and initialization methods, respectively. Our goal is to determine whether or not there is at least one method that is significantly better than at least one other method at the $\alpha = 0.05$ level. If this is the case, we will conduct a post-hoc (multiple comparison) test to determine which pairs of methods differ significantly. For this purpose, we will use the Bergmann-Hommel test \cite{Bergmann88}, a powerful post-hoc procedure that has been used successfully in various machine learning studies \cite{Garcia08, Voss10}.

\subsection{Real Data Sets}
\label{sec_real_data}

Table \ref{tab_real_ranking} gives the Final \texttt{SSE} rankings of the IMs for the real data sets as determined by the Bergmann-Hommel procedure using data given in Table \ref{tab_real_final_sse}. Here, a notation such as $\mathsf{C < \{D, E\}}$ indicates that there is no statistically significant difference between methods \textsf{D} and \textsf{E} and these two methods are significantly better than method \textsf{C}. From Table \ref{tab_real_ranking} it can be seen that the methods cannot be distinguished from each other reliably. This was expected since even nonparametric post-hoc tests lack discrimination power in small sample cases (recall that only $32$ data sets were used) with a large number of ties (see Table \ref{tab_real_final_sse}). For example, with respect to the \emph{minimum} statistic, the performances of \textsf{F}, \textsf{M}, \textsf{B}, \textsf{K}, and \textsf{G} are statistically indistinguishable. In other words, if we initialize k-means with each of these non-deterministic methods and execute it until convergence, the resulting clusterings over $R=100$ runs will have very similar \emph{minimum} Final \texttt{SSE} values. Similar trends were observed for the \texttt{RAND}, \texttt{VD}, and \texttt{VI} criteria (data not shown). Given the abundance of local minima even in data sets of moderate size and/or dimensionality and the gradient descent nature of k-means, it is not surprising that the deterministic methods \textsf{V} and \textsf{P} are outperformed by most of the non-deterministic methods as the former methods were executed only once, whereas the latter ones were executed $R=100$ times.
\par
As mentioned earlier, the \emph{minimum} statistic is meaningful only when it is practical to execute k-means multiple times. Otherwise, the \emph{mean} statistic is more meaningful. The analysis of \emph{mean} Final \texttt{SSE} results using the Bergmann-Hommel procedure reveals that deterministic methods \textsf{V} and \textsf{P} are the preferred choices in this case. This is not surprising since non-deterministic methods, in particular those that are \emph{ad hoc} in nature, often produce highly variable results across multiple runs.
\par
The \emph{standard deviation} statistic characterizes the reliability of a non-deterministic IM with respect to a particular performance criterion. In other words, if a non-deterministic IM obtains low \emph{mean} and \emph{standard deviation} with respect to an effectiveness criterion, we do not have to execute this method a large number of times to obtain good results. The analysis of Final \texttt{SSE} \emph{standard deviations} reveals two overlapping groups of methods. Once again this is not necessarily because the members of each group are in fact indistinguishable with respect to their reliability, but due to the relatively small sample size used. In summary, due to the necessarily small number of real-world data sets available for clustering studies, it may not be possible to distinguish among various IMs. Therefore, it is crucial that these IMs be tested on a large number of synthetic data sets (see \S\ref{sec_synt_data}).
\par
As for computational efficiency, it can be seen from Table \ref{tab_real_cpu_time} that, in general, the IMs have similar computational requirements per run. However, in practice, a non-deterministic method is typically executed $R$ times and the output of the run that gives the least \texttt{SSE} is taken as the result. Therefore, the total computational cost of a non-deterministic method is often significantly higher than that of a deterministic method. As predicted in \S \ref{sub_sec_linear_vs_super}, simple methods such as \texttt{M} require about the same CPU time as elaborate methods such as \texttt{G}. This is because simple methods often lead to more k-means iterations, whereas elaborate ones compensate for their computational overhead by requiring fewer k-means iterations. It should be noted that efficiency differences among the methods can be further reduced by using faster k-means variants such as those described in \cite{Kanungo02, Hamerly10}.

\subsection{Synthetic Data Sets}
\label{sec_synt_data}
Table \ref{tab_min_ranking} gives the ranking of the IMs with respect to the \emph{minimum} statistic. It can be seen that despite variations in rankings across the performance criteria, some general trends emerge:

\begin{itemize}
\renewcommand{\labelitemi}{$\triangleright$}
  \item Non-deterministic methods outperform the deterministic ones, i.e., \textsf{V} and \textsf{P}, except in the case of Initial \texttt{SSE}. As explained in \S\ref{sec_real_data}, this is due to the fact that the non-deterministic methods were executed $R=100$ times, whereas the deterministic ones were executed only once. The reason why deterministic methods have good Initial \texttt{SSE} performance is because these methods are approximate (divisive hierarchical) clustering methods by themselves and thus they give reasonable results even without k-means refinement.
  \item Method \textsf{B} consistently appears in the best performing group, whereas methods \textsf{F} and \textsf{X} are often among the worst non-deterministic methods.
  \item Method \textsf{M} exhibits moderate-to-good performance except in the case of Initial \texttt{SSE}. Recall that this method randomly selects the $K$ initial centers from among the data points and therefore it cannot be expected to perform well without k-means refinement.
  \item Methods \textsf{K} and \textsf{G} generally perform well. In some cases the latter outperforms the former, whereas in others they have comparable performance.
\end{itemize}

Table \ref{tab_mean_ranking} gives the ranking of the IMs with respect to the \emph{mean} statistic. It can be seen that despite variations in rankings across the performance criteria, some general trends emerge:

\begin{itemize}
\renewcommand{\labelitemi}{$\triangleright$}
  \item Deterministic methods, i.e., \textsf{V} and \textsf{P}, generally outperform the non-deterministic ones. As explained in \S\ref{sec_real_data}, this is due to the fact that the non-deterministic methods can produce highly variable results across multiple runs. Method \textsf{B} is highly competitive with the deterministic methods.
  \item Methods \textsf{M} and \textsf{X} are often among the worst performers, whereas methods \textsf{F} and \textsf{K} exhibit moderate-to-bad performance.
  \item Method \textsf{G} is often significantly better than all non-deterministic methods but \textsf{B}.
\end{itemize}

Table \ref{tab_stdev_ranking} gives the ranking of the non-deterministic IMs with respect to the \emph{standard deviation} statistic. It can be seen that despite variations in rankings across the performance criteria, some general trends emerge:

\begin{itemize}
\renewcommand{\labelitemi}{$\triangleright$}
  \item Method \textsf{B} consistently appears in the best performing group, whereas method \textsf{M} is often among the worst performers.
  \item Methods \textsf{X} and \textsf{K} exhibit moderate-to-bad performance.
  \item Method \textsf{F} and \textsf{G} are significantly better than all methods but \textsf{B}.
\end{itemize}

\subsection{Recommendations for Practitioners}

Based on the statistical analyses presented in the previous section, the following recommendations can be made:
\begin{itemize}
\renewcommand{\labelitemi}{$\triangleright$}
	\item In general, methods \textsf{F}, \textsf{M}, and \textsf{X} should not be used. These methods are easy to understand and implement, but they are often ineffective and unreliable. Furthermore, despite their low overhead, these methods do not offer significant time savings since they often result in slower k-means convergence.
	\item In time-critical applications that involve large data sets or applications that demand determinism, methods \textsf{V} or \textsf{P} should be used. These methods need to be executed only once and they lead to very fast k-means converge. The efficiency difference between the two is noticeable only on high dimensional data sets. This is because method \textsf{V} calculates the direction of split by determining the coordinate axis with the greatest variance (in $\mathcal{O}(D)$ time), whereas method \textsf{P} achieves this by calculating the principal eigenvector of the covariance matrix (in $\mathcal{O}(D^2)$ time using the power method \cite{Hotelling36}). Note that despite its higher computational complexity, method \textsf{P} can, in some cases, be more efficient than method \textsf{V} (see Table \ref{tab_real_cpu_time}). This is because the former converges significantly faster than the latter (see Table \ref{tab_mean_ranking}). The main disadvantage of these methods is that they are more complicated to implement due to their hierarchical formulation.
	\item In applications that involve small data sets, e.g., $N < 10,000$, methods \textsf{B} or \textsf{G} should be used. It is computationally feasible to run these methods hundreds of times on such data sets given that one such run takes only a few milliseconds.
	\item In applications where an approximate clustering of the data set is desired, methods \textsf{B}, \textsf{G}, \textsf{V}, or \textsf{P} should be used. These methods produce very good initial clusterings (see Tables \ref{tab_min_ranking} and \ref{tab_mean_ranking}), which makes it possible to use them as standalone clustering algorithms.
\end{itemize}

\section{Conclusions}
\label{sec_conc}
In this paper we presented an overview of k-means initialization methods with an emphasis on their computational efficiency. We then compared eight commonly used linear time initialization methods on a large and diverse collection of real and synthetic data sets using various performance criteria. Finally, we analyzed the experimental results using non-parametric statistical tests and provided recommendations for practitioners. Our statistical analyses revealed that popular initialization methods such as \emph{forgy}, \emph{macqueen}, and \emph{maximin} often perform poorly and that there are significantly better alternatives to these methods that have comparable computational requirements.

\section{Acknowledgments}

This publication was made possible by grants from the Louisiana Board of Regents (LEQSF2008-11-RD-A-12) and National Science Foundation (0959583, 1117457). The authors are grateful to D.\ Arthur, S.\ Vassilvitskii, J.G.\ Dy, S.J.\ Redmond, J.F.\ Lu, and M.\ Al Hasan for clarifying various points about their papers.

\bibliographystyle{elsarticle-num}

\newpage

{
\scriptsize
\begin{longtable}{@{\extracolsep{\fill}}lccccccccc}
\caption{\label{tab_real_final_sse} Final \texttt{SSE} (Real Data Sets)}\\
\hline
& & \textsf{F} & \textsf{M} & \textsf{X} & \textsf{B} & \textsf{K} & \textsf{G} & \textsf{V} & \textsf{P}\\
\hline
\hline
\multirow{2}{*}{1} & min & 239 & 239 & 239 & 239 & 239 & 239 & 239 & 239\\
& mean & 239$\pm$0 & 239$\pm$0 & 239$\pm$0 & 239$\pm$0 & 239$\pm$0 & 239$\pm$0 & 239$\pm$0 & 239$\pm$0\\
\hline
\multirow{2}{*}{2} & min & 38 & 38 & 41 & 38 & 38 & 38 & 39 & 38\\
& mean & 44$\pm$4 & 40$\pm$1 & 45$\pm$3 & 39$\pm$1 & 39$\pm$1 & 39$\pm$1 & 39$\pm$0 & 38$\pm$0\\
\hline
\multirow{2}{*}{3} & min & 167 & 167 & 167 & 167 & 167 & 167 & 173 & 167\\
& mean & 176$\pm$8 & 176$\pm$7 & 172$\pm$6 & 171$\pm$4 & 174$\pm$5 & 173$\pm$4 & 173$\pm$0 & 167$\pm$0\\
\hline
\multirow{2}{*}{4} & min & 10057 & 10057 & 10057 & 10057 & 10058 & 10058 & 10101 & 10100\\
& mean & 10115$\pm$79 & 10080$\pm$20 & 10077$\pm$15 & 10076$\pm$18 & 10083$\pm$23 & 10080$\pm$18 & 10101$\pm$0 & 10100$\pm$0\\
\hline
\multirow{2}{*}{5} & min & 66224 & 66224 & 66224 & 66224 & 66224 & 66224 & 66238 & 66238\\
& mean & 66990$\pm$890 & 67196$\pm$1048 & 67350$\pm$834 & 66431$\pm$360 & 66948$\pm$876 & 66930$\pm$773 & 66238$\pm$0 & 66238$\pm$0\\
\hline
\multirow{2}{*}{6} & min & 17 & 17 & 19 & 17 & 17 & 17 & 17 & 18\\
& mean & 19$\pm$2 & 19$\pm$3 & 20$\pm$1 & 18$\pm$1 & 18$\pm$1 & 18$\pm$1 & 17$\pm$0 & 18$\pm$0\\
\hline
\multirow{2}{*}{7} & min & 1167 & 1167 & 1267 & 1167 & 1167 & 1167 & 1167 & 1168\\
& mean & 1231$\pm$83 & 1250$\pm$103 & 1303$\pm$53 & 1184$\pm$25 & 1230$\pm$61 & 1198$\pm$35 & 1167$\pm$0 & 1168$\pm$0\\
\hline
\multirow{2}{*}{8} & min & 18 & 18 & 19 & 18 & 18 & 18 & 19 & 19\\
& mean & 20$\pm$1 & 20$\pm$2 & 22$\pm$2 & 20$\pm$1 & 20$\pm$2 & 20$\pm$1 & 19$\pm$0 & 19$\pm$0\\
\hline
\multirow{2}{*}{9} & min & 243 & 243 & 243 & 243 & 243 & 243 & 248 & 243\\
& mean & 251$\pm$8 & 252$\pm$8 & 253$\pm$8 & 251$\pm$7 & 252$\pm$8 & 249$\pm$7 & 248$\pm$0 & 243$\pm$0\\
\hline
\multirow{2}{*}{10} & min & 629 & 629 & 629 & 629 & 629 & 629 & 629 & 629\\
& mean & 629$\pm$0 & 633$\pm$28 & 671$\pm$81 & 637$\pm$39 & 635$\pm$34 & 635$\pm$35 & 629$\pm$0 & 629$\pm$0\\
\hline
\multirow{2}{*}{11} & min & 117891 & 117924 & 120898 & 117863 & 117719 & 117995 & 118495 & 118386\\
& mean & 119931$\pm$1060 & 119655$\pm$1061 & 123388$\pm$1264 & 119050$\pm$699 & 119538$\pm$894 & 119176$\pm$710 & 118495$\pm$0 & 118386$\pm$0\\
\hline
\multirow{2}{*}{12} & min & 1742 & 1742 & 1742 & 1742 & 1742 & 1742 & 1742 & 1742\\
& mean & 1742$\pm$0 & 1742$\pm$0 & 1742$\pm$0 & 1742$\pm$0 & 1744$\pm$28 & 1747$\pm$41 & 1742$\pm$0 & 1742$\pm$0\\
\hline
\multirow{2}{*}{13} & min & 2723 & 2718 & 2721 & 2719 & 2718 & 2715 & 2735 & 2745\\
& mean & 2775$\pm$28 & 2756$\pm$22 & 2765$\pm$17 & 2742$\pm$15 & 2754$\pm$18 & 2752$\pm$17 & 2735$\pm$0 & 2745$\pm$0\\
\hline
\multirow{2}{*}{14} & min & 2923 & 2923 & 2923 & 2923 & 2923 & 2923 & 2923 & 2923\\
& mean & 2923$\pm$0 & 2923$\pm$0 & 2923$\pm$0 & 2923$\pm$0 & 2923$\pm$0 & 2923$\pm$0 & 2923$\pm$0 & 2923$\pm$0\\
\hline
\multirow{2}{*}{15} & min & 3127 & 3128 & 3180 & 3127 & 3128 & 3127 & 3137 & 3214\\
& mean & 3164$\pm$30 & 3168$\pm$28 & 3247$\pm$22 & 3157$\pm$29 & 3173$\pm$33 & 3149$\pm$20 & 3137$\pm$0 & 3214$\pm$0\\
\hline
\multirow{2}{*}{16} & min & 2802 & 2802 & 2802 & 2802 & 2802 & 2802 & 21983 & 2802\\
& mean & 8229$\pm$8685 & 12518$\pm$9667 & 2802$\pm$0 & 11935$\pm$9656 & 5722$\pm$6944 & 3774$\pm$4236 & 21983$\pm$0 & 2802$\pm$0\\
\hline
\multirow{2}{*}{17} & min & 36373 & 36373 & 36373 & 36373 & 36373 & 36373 & 36373 & 36373\\
& mean & 37755$\pm$2829 & 37046$\pm$916 & 36738$\pm$754 & 37152$\pm$1340 & 37440$\pm$1906 & 37103$\pm$1639 & 36373$\pm$0 & 36373$\pm$0\\
\hline
\multirow{2}{*}{18} & min & 14559 & 14559 & 14559 & 14559 & 14559 & 14559 & 14581 & 14807\\
& mean & 14653$\pm$140 & 14763$\pm$273 & 14774$\pm$293 & 14627$\pm$66 & 14735$\pm$234 & 14719$\pm$214 & 14581$\pm$0 & 14807$\pm$0\\
\hline
\multirow{2}{*}{19} & min & 215 & 215 & 230 & 215 & 215 & 215 & 227 & 215\\
& mean & 217$\pm$4 & 217$\pm$4 & 254$\pm$32 & 219$\pm$6 & 219$\pm$10 & 217$\pm$4 & 227$\pm$0 & 215$\pm$0\\
\hline
\multirow{2}{*}{20} & min & 235 & 219 & 233 & 218 & 217 & 217 & 220 & 219\\
& mean & 251$\pm$7 & 224$\pm$2 & 241$\pm$3 & 222$\pm$2 & 220$\pm$2 & 219$\pm$1 & 220$\pm$0 & 219$\pm$0\\
\hline
\multirow{2}{*}{21} & min & 4930 & 4930 & 4930 & 4930 & 4930 & 4930 & 4930 & 5004\\
& mean & 5130$\pm$131 & 5091$\pm$110 & 5036$\pm$106 & 5012$\pm$70 & 5111$\pm$116 & 5046$\pm$75 & 4930$\pm$0 & 5004$\pm$0\\
\hline
\multirow{2}{*}{22} & min & 1177 & 1177 & 1195 & 1177 & 1177 & 1177 & 1182 & 1177\\
& mean & 1179$\pm$10 & 1187$\pm$18 & 1204$\pm$25 & 1182$\pm$12 & 1193$\pm$27 & 1183$\pm$14 & 1182$\pm$0 & 1177$\pm$0\\
\hline
\multirow{2}{*}{23} & min & 121 & 121 & 121 & 121 & 121 & 121 & 121 & 121\\
& mean & 121$\pm$2 & 122$\pm$5 & 122$\pm$3 & 122$\pm$3 & 122$\pm$5 & 122$\pm$5 & 121$\pm$0 & 121$\pm$0\\
\hline
\multirow{2}{*}{24} & min & 387 & 387 & 411 & 387 & 387 & 387 & 410 & 405\\
& mean & 407$\pm$23 & 414$\pm$20 & 430$\pm$21 & 402$\pm$16 & 410$\pm$19 & 402$\pm$13 & 410$\pm$0 & 405$\pm$0\\
\hline
\multirow{2}{*}{25} & min & 235 & 235 & 411 & 235 & 235 & 235 & 235 & 274\\
& mean & 307$\pm$39 & 275$\pm$23 & 930$\pm$105 & 244$\pm$18 & 271$\pm$39 & 246$\pm$21 & 235$\pm$0 & 274$\pm$0\\
\hline
\multirow{2}{*}{26} & min & 214 & 214 & 214 & 214 & 214 & 214 & 214 & 214\\
& mean & 214$\pm$0 & 214$\pm$0 & 214$\pm$0 & 214$\pm$0 & 214$\pm$0 & 214$\pm$0 & 214$\pm$0 & 214$\pm$0\\
\hline
\multirow{2}{*}{27} & min & 22 & 22 & 22 & 22 & 22 & 22 & 23 & 23\\
& mean & 23$\pm$2 & 23$\pm$1 & 23$\pm$0 & 23$\pm$1 & 23$\pm$1 & 23$\pm$0 & 23$\pm$0 & 23$\pm$0\\
\hline
\multirow{2}{*}{28} & min & 223 & 223 & 224 & 223 & 223 & 223 & 224 & 224\\
& mean & 224$\pm$2 & 226$\pm$4 & 237$\pm$1 & 228$\pm$6 & 226$\pm$5 & 225$\pm$3 & 224$\pm$0 & 224$\pm$0\\
\hline
\multirow{2}{*}{29} & min & 7772 & 7772 & 7772 & 7772 & 7772 & 7772 & 7774 & 7774\\
& mean & 7798$\pm$91 & 7808$\pm$102 & 7854$\pm$160 & 7773$\pm$1 & 7831$\pm$140 & 7811$\pm$106 & 7774$\pm$0 & 7774$\pm$0\\
\hline
\multirow{2}{*}{30} & min & 334 & 334 & 348 & 334 & 334 & 334 & 335 & 334\\
& mean & 335$\pm$2 & 336$\pm$2 & 374$\pm$17 & 337$\pm$5 & 336$\pm$3 & 336$\pm$3 & 335$\pm$0 & 334$\pm$0\\
\hline
\multirow{2}{*}{31} & min & 11039 & 11039 & 11039 & 11039 & 11039 & 11039 & 11483 & 12422\\
& mean & 14041$\pm$1686 & 12367$\pm$1057 & 11714$\pm$627 & 11128$\pm$231 & 11773$\pm$872 & 11493$\pm$626 & 11483$\pm$0 & 12422$\pm$0\\
\hline
\multirow{2}{*}{32} & min & 58 & 58 & 61 & 58 & 58 & 58 & 69 & 59\\
& mean & 64$\pm$5 & 70$\pm$6 & 61$\pm$1 & 66$\pm$6 & 63$\pm$5 & 59$\pm$1 & 69$\pm$0 & 59$\pm$0\\
\hline
\end{longtable}
}

{
\scriptsize
\begin{longtable}{@{\extracolsep{\fill}}lrrrrrrrr}
\caption{\label{tab_real_cpu_time} CPU Time (Real Data Sets)}\\
\hline
& \textsf{F} & \textsf{M} & \textsf{X} & \textsf{B} & \textsf{K} & \textsf{G} & \textsf{V} & \textsf{P}\\
\hline
\hline
1 & 0 & 0 & 0 & 0 & 0 & 0 & 0 & 0\\
\hline
2 & 3 & 3 & 2 & 4 & 3 & 2 & 10 & 0\\
\hline
3 & 2 & 2 & 2 & 4 & 2 & 2 & 0 & 10\\
\hline
4 & 2295 & 2248 & 2173 & 3624 & 2332 & 2459 & 1900 & 2540\\
\hline
5 & 2183 & 2229 & 2714 & 3604 & 2273 & 2274 & 1730 & 2120\\
\hline
6 & 0 & 0 & 0 & 1 & 0 & 0 & 0 & 0\\
\hline
7 & 8 & 8 & 7 & 12 & 9 & 9 & 0 & 20\\
\hline
8 & 0 & 0 & 0 & 1 & 0 & 0 & 0 & 0\\
\hline
9 & 0 & 1 & 0 & 1 & 0 & 1 & 0 & 0\\
\hline
10 & 0 & 0 & 0 & 1 & 0 & 0 & 0 & 0\\
\hline
11 & 3730 & 3469 & 2469 & 4063 & 3537 & 3915 & 6940 & 12200\\
\hline
12 & 28 & 32 & 40 & 40 & 34 & 32 & 40 & 50\\
\hline
13 & 700 & 693 & 729 & 852 & 698 & 693 & 950 & 800\\
\hline
14 & 20 & 19 & 28 & 30 & 21 & 19 & 30 & 20\\
\hline
15 & 54 & 56 & 62 & 70 & 57 & 58 & 30 & 70\\
\hline
16 & 252 & 283 & 58 & 417 & 112 & 96 & 230 & 570\\
\hline
17 & 22 & 20 & 24 & 34 & 25 & 26 & 20 & 220\\
\hline
18 & 112 & 116 & 131 & 137 & 121 & 125 & 60 & 140\\
\hline
19 & 9 & 10 & 7 & 12 & 8 & 10 & 10 & 10\\
\hline
20 & 109 & 100 & 110 & 150 & 97 & 118 & 60 & 70\\
\hline
21 & 59 & 53 & 52 & 67 & 56 & 59 & 30 & 50\\
\hline
22 & 430 & 524 & 314 & 718 & 469 & 513 & 730 & 480\\
\hline
23 & 1 & 1 & 1 & 1 & 1 & 0 & 0 & 0\\
\hline
24 & 6 & 5 & 7 & 8 & 6 & 7 & 10 & 0\\
\hline
25 & 84 & 82 & 19 & 122 & 79 & 81 & 100 & 80\\
\hline
26 & 0 & 0 & 0 & 1 & 0 & 0 & 10 & 0\\
\hline
27 & 1 & 1 & 0 & 1 & 1 & 1 & 0 & 0\\
\hline
28 & 3 & 2 & 1 & 3 & 2 & 3 & 10 & 10\\
\hline
29 & 20 & 20 & 20 & 28 & 19 & 21 & 10 & 20\\
\hline
30 & 38 & 38 & 24 & 51 & 36 & 38 & 60 & 40\\
\hline
31 & 748 & 949 & 1377 & 2439 & 918 & 1044 & 580 & 840\\
\hline
32 & 5 & 6 & 5 & 9 & 6 & 6 & 0 & 0\\
\hline
\end{longtable}
}

\begin{table}[ht]
\centering
\linespread{1}
\caption{ \label{tab_real_ranking} Final \texttt{SSE} Rankings (Real Data Sets)}
\begin{tabular}{lc}
\hline
Statistic & IM Ranking\\
\hline
\hline
Minimum & $\mathsf{\{X, V, P\} < \{F, M, B, K, G\}}$\\
\hline
Mean & \textcolor{black}{$\mathsf{\{F, M, X, K\} < \{F, B, G\} < \{G, V\} < \{V, P\}}$}\\
\hline
Standard Deviation & \textcolor{black}{$\mathsf{\{F, M, X, B, K\} < \{X, B, G\}}$}\\
\end{tabular}
\end{table}

\begin{table}[ht]
\centering
\linespread{1}
\caption{ \label{tab_min_ranking} Minimum (Synthetic Data Sets)}
\begin{tabular}{lc}
\hline
Data Set Complexity & IM Ranking\\
\hline
\hline
\multicolumn{2}{c}{Initial \texttt{SSE}}\\
\hline
Easy & $\mathsf{X < \{F, M\} < K < G < V < P < B}$\\
Moderate & $\mathsf{X < \{F, M, K\} < G < V < P < B}$\\
Difficult & $\mathsf{X < \{M, K\} < F < G < V < P < B}$\\
\hline
\multicolumn{2}{c}{Final \texttt{SSE}}\\
\hline
Easy & $\mathsf{\{V, P\} < \{F, X\} < \{M, B, K, G\}}$\\
Moderate & \textcolor{black}{$\mathsf{\{V, P\} < \{F, X\} < \{F, M, B, K\} < \{M, B, K, G\}}$}\\
Difficult & \textcolor{black}{$\mathsf{V < P < \{F, X\} < \{M, X, B, K, G\}}$}\\
\hline
\multicolumn{2}{c}{Final \texttt{RAND}}\\
\hline
Easy & $\mathsf{\{V, P\} < X < F < \{M, K\} < G < B}$\\
Moderate & \textcolor{black}{$\mathsf{\{V, P\} < X < \{F, M, K\} < \{M, K, G\} < \{B, G\}}$}\\
Difficult & \textcolor{black}{$\mathsf{V < P < \{F, X\} < \{F, K, G\} < \{M, K, G\} < \{M, B, K\}}$}\\
\hline
\multicolumn{2}{c}{Final \texttt{VD}}\\
\hline
Easy & $\mathsf{\{V, P\} < X < \{F, M\} <  \{K, G\} < B}$\\
Moderate & \textcolor{black}{$\mathsf{\{V, P\} < X < \{F, M\} < \{M, K\} < \{B, K, G\}}$}\\
Difficult & \textcolor{black}{$\mathsf{V < P < \{F, X, G\} < \{M, K, G\} < \{M, B, K\}}$}\\
\hline
\multicolumn{2}{c}{Final \texttt{VI}}\\
\hline
Easy & $\mathsf{\{V, P\} < X < \{F, M, K, G\} < B}$\\
Moderate & $\mathsf{\{V, P\} < X < F < \{M, B, K, G\}}$\\
Difficult & \textcolor{black}{$\mathsf{V < P < \{F, X\} < \{X, G\} < \{M, B, K, G\}}$}\\
\hline
\multicolumn{2}{c}{Number of Iterations}\\
\hline
Easy & $\mathsf{V < F < M < \{X, K\} < P < G < B}$\\
Moderate & $\mathsf{V < P < F < M < K < \{X, G\} < B}$\\
Difficult & $\mathsf{V < P < F < M < K < G < X < B}$\\
\hline
\end{tabular}
\end{table}

\begin{table}[ht]
\centering
\linespread{1}
\caption{ \label{tab_mean_ranking} Mean (Synthetic Data Sets)}
\begin{tabular}{lc}
\hline
Data Set Complexity & IM Ranking\\
\hline
\hline
\multicolumn{2}{c}{Initial \texttt{SSE}}\\
\hline
Easy & $\mathsf{X < M < K < F < G < V < \{B, P\}}$\\
Moderate & $\mathsf{X < \{M, K\} < F < G < V < \{B, P\}}$\\
Difficult & $\mathsf{X < K < M < F < G < V < B < P}$\\
\hline
\multicolumn{2}{c}{Final \texttt{SSE}}\\
\hline
Easy & $\mathsf{\{M, X\} < K < F < G < B < \{V, P\}}$\\
Moderate & $\mathsf{X < \{M, K\} < F < G < B < V < P}$\\
Difficult & $\mathsf{F < \{M, K\} < X < G < B < V < P}$\\
\hline
\multicolumn{2}{c}{Final \texttt{RAND}}\\
\hline
Easy & $\mathsf{\{M, X\} < K < F < G < B < \{V, P\}}$\\
Moderate & $\mathsf{X < \{M, K\} < \{F, G\} < B < \{V, P\}}$\\
Difficult & \textcolor{black}{$\mathsf{\{F, K\} < \{X, K, G\} < M < V < B < P}$}\\
\hline
\multicolumn{2}{c}{Final \texttt{VD}}\\
\hline
Easy & $\mathsf{\{M, X\} < K < F < G < B < \{V, P\}}$\\
Moderate & $\mathsf{X < \{M, K\} < F < G < B < V < P}$\\
Difficult & $\mathsf{F < \{M, X, K, G\} < V < B < P }$\\
\hline
\multicolumn{2}{c}{Final \texttt{VI}}\\
\hline
Easy & $\mathsf{\{M, X\} < K < F < G < B < \{V, P\}}$\\
Moderate & $\mathsf{\{M, X, K\} < F < G < B < V < P}$\\
Difficult & \textcolor{black}{$\mathsf{F < \{M, K, G\} < \{X, G\} < V < B < P}$}\\
\hline
\multicolumn{2}{c}{Number of Iterations}\\
\hline
Easy & $\mathsf{M < X < K < F < G < V < P < B}$\\
Moderate & $\mathsf{\{F, M\} < \{X, K\} < G < V < P < B}$\\
Difficult & $\mathsf{F < M < K < G < X < V < P < B}$\\
\hline
\end{tabular}
\end{table}

\begin{table}[ht]
\centering
\linespread{1}
\caption{ \label{tab_stdev_ranking} Standard Deviation (Synthetic Data Sets)}
\begin{tabular}{lc}
\hline
Data Set Complexity & IM Ranking\\
\hline
\hline
\multicolumn{2}{c}{Initial \texttt{SSE}}\\
\hline
Easy & $\mathsf{X < M < K < G < F < B}$\\
Moderate & Same as Easy\\
Difficult & $\mathsf{X < M < K < G < \{F, B\}}$\\
\hline
\multicolumn{2}{c}{Final \texttt{SSE}}\\
\hline
Easy & $\mathsf{M < \{X, K\} < G < F < B}$\\
Moderate & $\mathsf{\{M, K\} < X < \{F, G\} < B}$\\
Difficult & $\mathsf{\{M, K\} < \{F, X, G\} < B}$\\
\hline
\multicolumn{2}{c}{Final \texttt{RAND}}\\
\hline
Easy & $\mathsf{M < \{X, K\} < G < F < B}$\\
Moderate & $\mathsf{\{M, X, K\} < \{F, G\} < B}$\\
Difficult & $\mathsf{\{M, K\} < \{F, X, G\} < B}$\\
\hline
\multicolumn{2}{c}{Final \texttt{VD}}\\
\hline
Easy & $\mathsf{M < \{X, K\} < G < F < B}$\\
Moderate & $\mathsf{\{M, K\} < X < \{F, G\} < B}$\\
Difficult & \textcolor{black}{$\mathsf{\{M, K\} < \{K, G\} < \{F, G\} < X < B}$}\\
\hline
\multicolumn{2}{c}{Final \texttt{VI}}\\
\hline
Easy & $\mathsf{M < \{X, K\} < \{F, G\} < B}$\\
Moderate & $\mathsf{\{M, K\} < X < \{F, G\} < B}$\\
Difficult & $\mathsf{\{F, M, K, G\} < X < B}$\\
\hline
\multicolumn{2}{c}{Number of Iterations}\\
\hline
Easy & $\mathsf{M < K < \{X, G\} < F < B}$\\
Moderate & $\mathsf{\{M, K\} < X < \{F, G\} < B}$\\
Difficult & $\mathsf{\{F, M, X, K, G\} < B}$\\
\hline
\end{tabular}
\end{table}

\end{document}